# Tuning structure learning algorithms with out-of-sample and resampling strategies

Kiattikun Chobtham  k.chobtham@qmul.ac.uk

Anthony C. Constantinou  a.constantinou@qmul.ac.uk

Bayesian Artificial Intelligence research lab, Machine Intelligence and Decision Systems (MInDS) research group, School of Electronic Engineering and Computer Science, Queen Mary University of London, London, UK, E1 4NS.

**Abstract**

One of the challenges practitioners face when applying structure learning algorithms to their data involves determining a set of hyperparameters; otherwise, a set of hyperparameter defaults is assumed. The optimal hyperparameter configuration often depends on multiple factors, including the size and density of the usually unknown underlying true graph, the sample size of the input data, and the structure learning algorithm. We propose a novel hyperparameter tuning method, called the Out-of-sample Tuning for Structure Learning (OTSL), that employs out-of-sample and resampling strategies to estimate the optimal hyperparameter configuration for structure learning, given the input data set and structure learning algorithm. Synthetic experiments show that employing OTSL as a means to tune the hyperparameters of hybrid and score-based structure learning algorithms leads to improvements in graphical accuracy compared to the state-of-the-art. We also illustrate the applicability of this approach to real datasets from different disciplines.

**Keywords:** *Bayesian networks, bootstrapping, causal discovery, hyperparameter optimisation, probabilistic graphical models.*

## 1. Introduction

A causal Bayesian Network (BN) is a probabilistic graphical model that enables decision makers to reason under uncertainty, particularly in complex systems that require answers to interventional and counterfactual questions (Pearl, 1988). It is represented by a Directed Acyclic Graph (DAG) which consists of nodes corresponding to random variables and arcs corresponding to causal relationships, where an arc $A_k \rightarrow A_j$ is viewed as $A_k$ being the direct cause of node $A_j$. The causal relationships are represented by a set of conditional probabilities $P(A_i|\text{parent}(A_i))$, where $\text{parent}(A_i)$ refers to the parent-set of $A_i$. The joint distribution over all nodes V entails all conditional probabilities as follows:

$$P(A_1, \ldots, A_V) = \prod_{i=1}^{V} P(A_i|\text{parent}(A_i))$$

Learning the structure of a BN is generally NP-hard, where the number of possible graphs grows super-exponential in the number of variables, making it computationally intractable. There are two main classes of unsupervised structure learning algorithms; namely constraint-based and score-based learning. A constraint-based algorithm typically relies on statistical Conditional Independence (CI) tests (see Section 2.1) to construct a graph skeleton, and then orientate some of the edges that make up the skeleton. On the other hand, a score-based algorithm involves a search method that traverses the space of possible graphs, and returns the graph with the highest objective score (see Section 2.2). Algorithms that combine both of these learning strategies are common in the literature, and they are referred to as hybrid learning algorithms.

There are hundreds of structure learning algorithms in the literature (Kitson et al., 2021). One of the most well-established constraint-based algorithms is PC-Stable by Colombo and Maathuis (2014), which is based on PC by Spirtes and Glymour (1991) but addresses PC's sensitivity to the order of the variables as read from data. PC-Stable returns a Partially DAG (PDAG) that contains directed and undirected edges, and which can sometimes be converted into a Completed PDAG (CPDAG) that represents a set of Markov equivalence DAGs that encode the same CI statements. For instance, a serial connection (A → B → C) and a divergence connection (A ← B → C) encode $A \perp C \mid B$, indicating that the direction of these edges cannot be determined purely from observational data. Score-based algorithms will often return a DAG or a CPDAG output. However, if a score-based algorithm employs a score-equivalent objective function, the DAG output will represent a random DAG of the highest scoring Markov equivalence class (or CPDAG).

Traversing the search-space of graphs is computationally expensive and hence, most score-based algorithms tend to be approximate. For example, the Hill-Climing (HC) algorithm by Heckerman et al. (1995) and the FGES algorithm by Ramsey (2015) rely on greedy heuristics. HC performs hill-climbing search in the space of DAGs, whereas FGES greedily searches the space of CPDAGs. Approximate solutions include hybrid algorithms such as the Max Min Hill-Climbing (MMHC) by Tsamardinos et al. (2006), and hybrid MCMC by Kuipers et al. (2022). Hybrid algorithms tend to involve a restrict phase and a maximisation phase, and MMHC is a widely-used example of this; i.e., it starts with constraint-based learning and uses CI tests to determine the restricted space of DAGs, followed by applying hill-climbing search to the reduced space of DAGs. Hybrid MCMC works in a similar manner and starts by creating a restricted search space using PC, followed by MCMC sampling in the node ordering space of DAGs.

An issue with these algorithms is that they come with a set of unoptimised hyperparameters. Because there is little guidance on how to choose these hyperparameters, most papers in the literature use these algorithms with either their hyperparameter defaults, or test them over a restricted set of different plausible hyperparameter values; a process that can be very time consuming. Hyperparameter tuning for structure learning algorithms can be divided into in-sample tuning and out-of-sample tuning methods, where the former utilises all available data and the latter uses a subset of the available data as a test data to tune hyperparameter configurations on data points that were not included in the training set. In-sample tuning approaches include the Stability Approach to Regularization Selection (StARS) by Liu et al. (2010), which optimises for model stability by selecting the hyperparameter configuration that generates the most stable learnt graphs over perturbations of the input data. Out-of-sample tuning approaches include the Out-of-sample Causal Tuning (OCT) by Biza et al. (2020, 2022), which performs cross-validation to identify the Markov Blankets (MBs) for each variable. The MB of a variable A represents a set of variables that make A independent of all other variables, and can serve as a feature selection method. Specifically, the MB of A includes the parents of A, its children, and the parents of its children. The OCT algorithm uses MBs to obtain a Random Forest model and optimises hyperparameters for predictive accuracy over test data. Experimental results showed that it performed well against the in-sample StARS approach discussed above.

This paper proposes a novel hyperparameter tuning method that employs out-of-sample and resampling strategies to estimate the optimal hyperparameter configuration for structure learning, given the input dataset and structure learning algorithm. The paper is organised as follows: Section 2 provides preliminary information on hyperparameters in the context of structure learning, Section 3 describes the proposed hyperparameter tuning approach, Section 4 presents the results, and we provide our concluding remarks in Section 5.

## 2. Preliminary Information

This section provides an overview of the common CI tests used by constraint-based algorithms, and the common objective functions used by score-based algorithms. Subsections 2.1 and 2.2 cover the hyperparameters that could be tuned for functions that test for CI and for objective functions respectively.

### 2.1. Functions that test for Conditional Independence (CI)

#### 2.1.1. Pearson's chi$^2$

The Pearson's chi$^2$ statistical test (Pearson, 1900) is a commonly used function for testing CI given discrete data. It assumes the null hypothesis that node A and node B are conditionally independent given the set of nodes **C**. The test produces a p-value of the test statistic which is used to determine whether to reject or accept the null hypothesis. The significance threshold ($\alpha$) serves as the hyperparameter of the Pearson's chi$^2$ test, and is set to 0.05 by convention. If the p-value is less than $\alpha$, the null hypothesis is rejected, and node A and node B are assumed to be conditionally dependent given **C**. If the p-value is greater than $\alpha$, the null hypothesis is not rejected and hence, node A and node B are assumed to be conditionally independent given **C**. The formula for the Pearson's chi$^2$ test is:

$$\chi^2 = 2 \sum \frac{(n_{abc} - m_{abc})^2}{n_{abc}}$$

where $n_{abc}$ is the number of instances in the data D where A = a, B = b and **C** = c, $m_{abc} = \frac{n_{ac} \cdot n_{bc}}{n_c}$, and the process of calculating the number of instances of $n_{ac}$, $n_{bc}$ and $n_c$ is analogous to that of $n_{abc}$. We will use Chi$^2$ interchangeably with the Pearson's chi$^2$ test for the rest of this paper.

#### 2.1.2. Mutual information (MI)

Shannon's Mutual Information (MI) was introduced as a measure of mutual dependence between two discrete variables (Cover and Thomas 2006). The MI between two nodes A and B is defined as:

$$MI(A, B) = \sum_{a,b} \hat{p}(a,b) \ln \left[ \frac{\hat{p}(a,b)}{\hat{p}(a)\hat{p}(b)} \right]$$

where $\hat{p}(a,b)$ refers to $\hat{p}(A = a, B = b)$ as the probability $p(a,b)$ derived from the maximum likelihood estimate. It is calculated as $\hat{p}(a,b) = \frac{n_{ab}}{n}$, where n is the total number of samples, and the process of calculating $\hat{p}(a)$ and $\hat{p}(b)$ is analogous to that of $\hat{p}(a,b)$. Consequently, conditional MI can be used for CI test, defined as:

$$MI(A, B \mid \mathbf{C}) = \sum_{a,b,c} \hat{p}(a,b,c) \ln \left[ \frac{\hat{p}(a,b,c)\hat{p}(c)}{\hat{p}(a,c) \cdot \hat{p}(b,c)} \right]$$

where $\hat{p}(a,b,c) = \frac{n_{abc}}{n}$, and the process of calculating $\hat{p}(a,c)$, $\hat{p}(b,c)$ and $\hat{p}(c)$ is analogous to that of $\hat{p}(a,b,c)$. The significance threshold $\alpha$ serves the same purposed as in the Chi$^2$ test, i.e., if MI(A, B | **C**) is greater than $\alpha$, node A and node B are conditionally independent given **C**.

#### 2.1.3. Shrinkage Mutual Information test (MI-sh)

James and Stein (1961) proposed a shrinkage estimate of MI for two random variables in the form of a regulariser, which they call the James-Stein-type shrinkage intensity λ. The conditional MI-sh test (Scutari and Brogini, 2012) between A and B given **C** is defined as the expectation of $MI - sh(A, B|\mathbf{C})$ with respect to the distribution of **C**. As with the Chi² and MI tests, they use the significance threshold α to accept or reject the same null hypothesis. The MI-sh test is described as follows:

$$MI - sh(A, B|\mathbf{C}) = \sum_{a,b,c} p^{shrink}(a, b, c) \log \left[ \frac{p^{shrink}(a, b, c) p^{shrink}(c)}{p^{shrink}(a, c) p^{shrink}(b, c)} \right]$$

where;

$$p^{shrink}(a, b, c) = \lambda \frac{1}{|A||B||\mathbf{C}|} + (1 - \lambda)\hat{p}(a, b, c)$$

$$p^{shrink}(a, c) = \lambda \frac{1}{|A||\mathbf{C}|} + (1 - \lambda)\hat{p}(a, c)$$

$$p^{shrink}(b, c) = \lambda \frac{1}{|B||\mathbf{C}|} + (1 - \lambda)\hat{p}(b, c)$$

$$p^{shrink}(c) = \lambda \frac{1}{|\mathbf{C}|} + (1 - \lambda)\hat{p}(c)$$

where |A|, |B| and |**C**| denote the number of states of variables A, B and the set of variables **C** respectively, and λ is the shrinkage intensity. Hausser and Strimmer (2008) proposed a closed-form estimator $\lambda^*$ that employs James-Stein-type shrinkage making it highly efficient computationally. In the case of estimating a single parameter, $\lambda^*$ is defined as:

$$\lambda^* = \frac{1 - \sum_{k=1}^{V} (\hat{p}_k)^2}{(n-1) \sum_{k=1}^{V} (\frac{1}{V} - \hat{p}_k)^2}$$

where $\lambda^* = [0,1]$ is the shrinkage intensity, $\lambda^* = 0$ means no shrinkage and $\lambda^* = 1$ full shrinkage, n is the sample size, and $\hat{p}_1 \ldots, \hat{p}_V$ are the probabilities of a given variable where $\sum_k \hat{p}_k = 1$.

*2.2. Objective functions*

*2.2.1. Bayesian Dirichlet with equivalent uniforms ($BDeu_{iss}$)*

$BDeu_{iss}$ is an objective function proposed by Heckerman et al. (1995) to determine the *maximum a posteriori* (MAP) structure by assuming equivalent uniform priors for a graph G. $BDeu_{iss}$ is score-equivalent, which means that it generates the same score for DAGs which entail the same joint probability distribution and are part of the same Markov equivalence class. Moreover, $BDeu_{iss}$ is a decomposable score where the score of a graph represents the sum of $BDeu_{iss}$ scores allocated to each node, given its parents, that is part of that graph. Decomposability is an important property that makes objective functions computationally efficient for structure learning. Traversing the search-space of graphs with a decomposable score implies that the scores of nodes that have not had a change to their parent-set can be carried over from the previous iteration, rather than re-computed.

Because the $BDeu_{iss}$ score is typically represented by a small value, its closed-form solution is expressed as a log function:

$$BDeu_{iss}(G, D) = \sum_{i=1}^{V} \sum_{j=1}^{q_i} \left[ \log \frac{\Gamma\left(iss/q_i\right)}{\Gamma\left(iss/q_i + N_{ij}\right)} + \sum_{k=1}^{r_i} \log \frac{\Gamma\left(iss/r_i q_i + N_{ijk}\right)}{\Gamma\left(iss/r_i q_i\right)} \right]$$

where G is a DAG, D is observational data, V is the number of variables, Γ is the Gamma function, $q_i$ denotes the number of possible combinations of values of parents of node $A_i$ and is equal to 1 if $A_i$ has no parents, j is the index of that combination, $N_{ijk}$ represents the number of instances in the data D where node $A_i$ has the $k^{th}$ value and its parents have the $j^{th}$ combination of values, $r_i$ is the number of states of node $A_i$, and $N_{ij} = \sum_{k=1}^{r_i} N_{ijk}$ represents the total number of instances in data D where parents of node $A_i$ have the $j^{th}$ combination of values. The *imaginary sample size* (iss) represents the hyperparameter of $BDeu_{iss}$ and it is often referred to the user's prior belief about the impact of the prior distribution on the objective score. Silander et al. (2006) observed that increasing iss led to a higher number of arcs and hence, denser learnt structures. They suggest that a reasonable iss for small sample sizes is between 1 to 20. In this work, we assume that the default iss hyperparameter for $BDeu_{iss}$ score is 1, as it is set in most studies and implementations of $BDeu_{iss}$.

### 2.2.1. Bayesian Information Criterion (BIC)

Schwarz (1978) proposed BIC as a model-selection function to reduce the risk of model-overfitting by balancing the goodness-of-fit with model dimensionality. It is based on Occam's razor principle in that the simplest solution is usually the best solution. Like $BDeu_{iss}$, BIC is decomposable and score equivalent. The general form of the score for discrete variables is expressed as:

$$BIC(G, D) = LL(G, D) - \frac{\log(n)}{2} F$$

where n is the sample size, $LL(G, D)$ denotes the Log-Likelihood (LL) of the data D given the graph G:

$$LL(G, D) = \log[\hat{p}(D|G)] = \sum_{i=1}^{V} \sum_{j=1}^{q_i} \sum_{k=1}^{r_i} N_{ijk} \log \frac{N_{ijk}}{N_{ij}}$$

and F is the complexity penalty represented by the number of free parameters of the model. It can be expressed as:

$$F = \sum_{i=1}^{V} (r_i - 1) q_i$$

Chen and Chen (2008) presented a modified version of BIC which they call Extended BIC (EBIC) that can be used to control the density of the learnt graph. This is achieved by introducing the hyperparameter $0 \leq \gamma$ that penalises the number of free parameters in the BN, which in turn are inversely proportional to the number of arcs in the learnt graph. This is equivalent to saying that large values of $\gamma$ will favour sparser graphs. EBIC is defined as:

$$EBIC_\gamma(G, D) = LL(G, D) - \frac{\log(n)}{2} F - \gamma \log(V) F, \qquad 0 \leq \gamma$$

Foygel and Drton (2010) studied the impact of the hyperparameter $\gamma' \in [0,1]$ and found that $\gamma' = 0.5$ is best in most synthetic experiments. However, it is acknowledged that the optimal value of $\gamma'$ is not invariant and hence, its optimisation remains an open question. In this paper, we define $EBIC_{normalised\ \gamma}$ as:

$$EBIC_{normalised\ \gamma}(G, D) = LL(G, D) - \frac{\log(n)}{2} F - \gamma' \log(V) F, \ \ 0 \leq \gamma' \leq 1$$

where the hyperparameter $0 \leq \gamma$ is normalised to $\gamma' \in [0,1]$. Thus, $\gamma$ is the hyperparameter of $EBIC_\gamma$ and $EBIC_{normalised\ \gamma}$ where $EBIC_{\gamma=0} = EBIC_{normalised\ \gamma=0} = BIC$.

## 3. Out-of-sample Tuning for Structure Learning (OTSL)

This section describes the algorithm we propose for hyperparameter tuning, which we call Out-of-sample Tuning for Structure Learning (OTSL). OTSL determines the optimal hyperparameter configuration for a structure learning algorithm by performing out-of-sample resampling and optimisation on test data.

### 3.1. Resampling with replacement with multiple training and test datasets

Resampling with replacement or bootstrapping (Efron and Tibshirani, 1994) is commonly used for sampling in statistics and machine learning. Unlike traditional cross-validation where each fold is drawn from a dataset without replacement, bootstrapping involves resampling with replacement to produce new data for validation that may contain multiple instances of the original cases. Resampling was previously applied to BN structure learning to address issues with data quality and was found to improve the accuracy of the learnt graph (Chun, 2011; Guo et al., 2022). We adopt this strategy for the OTSL algorithm and use resampling with replacement to generate multiple datasets for training and testing from a single observational dataset, where the training datasets are used for structure learning and the test datasets for hyperparameter tuning.

### 3.2. Tuning hyperparameters on test data

Section 2 describes both the BIC and $BDeu_{iss}$ scores, which are commonly used as objective functions in score-based structure learning algorithms. However, an issue with these model-selection scores is that the graph they score the highest tends not to be the ground truth graph. The model averaging MAHC by Constantinou et al. (2022) demonstrates that output graphs with slightly lower average BIC score may improve the graphical accuracy of the learnt graph, especially in the presence of data noise which is inevitably present in real data. This model averaging approach motivates the design of the proposed tuning approach, especially in that it focuses on maximising model selection by taking the average over multiple data splits.

We use the illustrations in Figure 1 to motivate our optimisation strategy, which is based on the HC algorithm and synthetic ALARM data with sample size 10k. Figure 1a presents the relationship between the graphical metric F1 (refer to Section 4) and the objective score $BDeu_{iss}$ when iss varies between 1 and 20. The tuning method involves resampling with replacement, where the input dataset of 10k is resampled 10 times and, at each iteration, split 9-to-1 for training and testing (refer to Algorithm 1). Specifically,

   i. $BDeu_{iss}$ is the tuning score optimised for different iss hyperparameters. Note that at each iteration of iss, the tuned score represents the average $BDeu_{iss}$ score over 10 iterations of resampling (refer to Algorithms 1 and 3).
   ii. F1 is the score for each graph recovered at different values of iss in $BDeu_{iss}$.

The illustration shows that it may be possible to optimise for iss in $BDeu_{iss}$ such that it improves the F1 score. Specifically, Figure 1a shows that the optimal value for iss in $BDeu_{iss}$ is 6, which in turn leads to a 0.57% improvement in F1 relative to the unoptimised hyperparameter default when iss = 1.

Figure 1b repeats the same exercise and assumes that the tuning score is $EBIC_{normalised\ \gamma}$, where $\gamma$ in $EBIC_\gamma$ varies between 0 and 19. In this example, we notice that the optimal γ hyperparameter is γ = 3 and happens to lead to the highest F1 score; an improvement of 11.63% relative to the unoptimised $EBIC_{normalised\ \gamma}$ score when γ = 0.

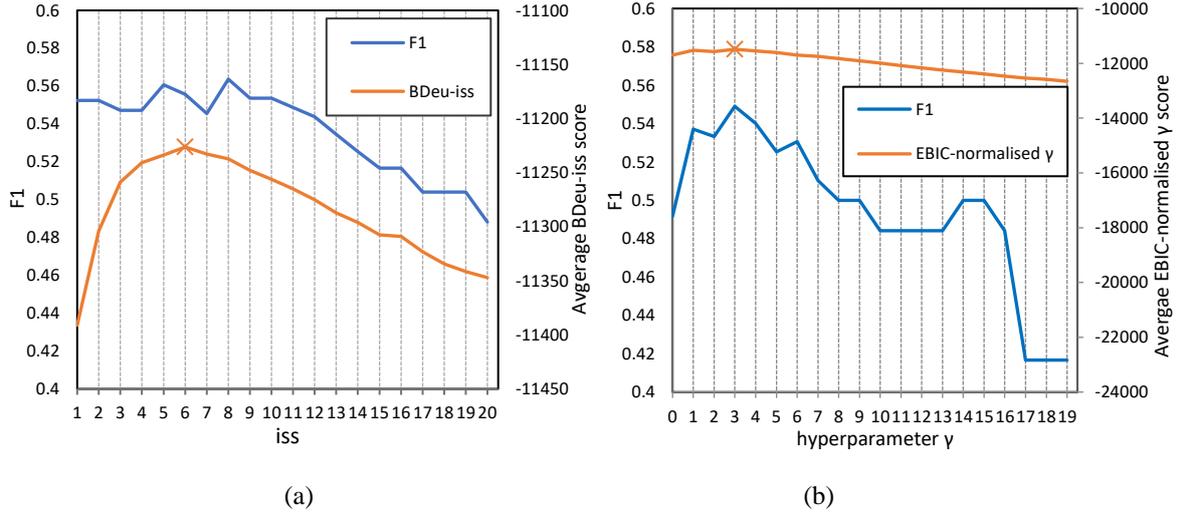

(a)            (b)

**Figure 1**: The F1 scores over different hyperparameter values for $BDeu_{iss}$ and $EBIC_\gamma$. The illustration is based on the HC algorithm and synthetic ALARM data with a sample size 10k.

*3.3. The Out-of-sample Tuning for Structure Learning (OTSL) algorithm*

Algorithm 1 describes the OTSL algorithm. As described in Algorithm 1, OTSL takes as input a dataset D, the number of iterations K for resampling (we assume 10 as default), the tuning score (we explore $BDeu_{iss}$ and $EBIC_{normalised\ \gamma}$ in this study) and a list of configurations C that specify the structure learning algorithm along with its hyperparameters and a range of those hyperparameters to be explored.

OTSL starts by resampling K training and test datasets given the input data. It then applies the specified structure learning algorithm with configurations C to each training dataset in K, and optimises the hyperparameters of either $BDeu_{iss}$ or $EBIC_{normalised\ \gamma}$ on each corresponding test dataset in K. The optimal configuration is the one that generates the highest average tuning score over K training and test datasets, and is returned as the optimal configuration. This process is described in Algorithms 1, 2 and 3, where Algorithms 2 and 3 describe the tuning process for $EBIC_{normalised\ \gamma}$ and $BDeu_{iss}$ respectively.

Algorithm 1: **O**ut-of-sample **T**uning for **S**tructure **L**earning (OTSL)

**Input:** dataset D, a list of configurations **C,** iteration K, **score for tuning**
**Output:** $c'$

1:    The sample size of train data X = the number of instances of D × (K-1) /K
2:    The sample size of test data Y = the number of instances of D / K
3:    **For** k = 1 to **K**
4:        $D_{k,training}$ ← **resample with replacement** (D) with sample size X
5:        $D_{k,test}$ ← **resample with replacement** (D \ $D_{k,training}$) with sample size Y
6:    **For** c ∈ **C** *// find the optimal configuration*
7:        **For** k = 1 to **K**
8:            $G_{c,k}$ ← **structure learning algorithm** ($D_{k,training}$, c)
9:            If $G_{c,k}$ is CPDAG
10:            $G_{c,k}$ ← **CPDAGtoDAG** ($G_{c,k}$)
11:            If $G_{c,k}$ is PDAG
12:            $G_{c,k}$ ← **PDAGtoDAG** ($G_{c,k}$)
13:            $S_{c,k}$ ← **score_for_tuning**($G_{c,k}$, $D_{k,test}$, c) // scoring functions given test data and hyperparameters
14:        $S_c$ = average $S_{c,k}$ over K
15:    $c'$ = arg max $S_c$
16:    return $c'$

---

Algorithm 2: **score_for_tuning** ($EBIC_{normalised\ \gamma}$)

**Input:** DAG G, dataset $\mathcal{D}$, configuration c from a list of configurations **C**
**Output:** $EBIC_{normalised\ \gamma}$

1:    **If** c contains γ
2:        score = $EBIC_{normalised\ \gamma}(G, \mathcal{D})$
3:    **Else**
4:        score = $EBIC_{normalised\ \gamma=0}(G, \mathcal{D})$

---

Algorithm 3: **score_for_tuning** ($BDeu_{iss}$)

**Input:** DAG G, dataset $\mathcal{D}$, configuration c from a list of configurations **C**
**Output:** $BDeu_{iss}$

1:    **If** c contains iss
2:        score = $BDeu_{iss}(G, \mathcal{D})$
3:    **Else**
4:        score = $BDeu_{iss=1}(G, \mathcal{D})$

## 4. Case studies and experimental setup

We consider 10 real-world BNs whose properties are provided in Table 1. Six of them are taken from the bnlearn (Scutari, 2019) and Bayesys (Constantinou et al., 2020) repositories and are used to generate synthetic data with sample sizes of 1k and 10k. In addition to the six synthetically generated datasets, we also consider four real datasets which we discuss in more detail in subsection 5.2.

| Synthetic data | Data source | Variables | Edges | Max in-degree | Free parameters |
|---|---|---|---|---|---|
| Asia | Bayesys (Constantinou et al., 2020) | 8 | 8 | 2 | 18 |
| Sports | | 9 | 15 | 2 | 1,049 |
| Property | | 27 | 31 | 3 | 3,056 |
| Alarm | | 37 | 46 | 4 | 509 |
| Hailfinder | bnlearn (Scutari, 2019) | 56 | 66 | 4 | 2,656 |
| Hepar2 | | 70 | 123 | 6 | 1,453 |
| **Real data** | **Data source** | **Variables** | **Sample size** | | |
| Diarrhoea | Bayesys (Constantinou et al., 2020) | 28 | 259,627 | | |
| Covid-19 | | 65 | 866 | | |
| ForMed | | 56 | 953 | | |
| Weather | NCEP (Kalnay et al., 1996) | 648 | 900 | | |

**Table 1.** The properties of the 10 case studies.

| Structure learning algorithm | Configuration | | | |
|---|---|---|---|---|
| | CI test / Objective function | Hyperparameter | | |
| | | α | γ | iss |
| **Constraint-based** | | | | |
| PC-Stable | Chi$^2$, MI, MI-sh | 0.01, [0.05], 0.1 | [0] | [1] |
| **Score-based** | | | | |
| HC, FGES | BDeu$_{iss}$, EBIC$_γ$ | - | [0], 1, 2, …, 19 | [1], 2, 3, …, 20 |
| **Hybrid based** | | | | |
| MCMC | Chi$^2$/ BDeu$_{iss}$ | [0.05] | - | [1], 2, 3, …, 20 |
| MMHC | Chi$^2$/ BDeu$_{iss}$, EBIC$_γ$ | 0.01, [0.05] | [0], 1, 2, …, 9 | [1], 2, 3, …, 10 |

**Table 2:** The algorithms tested for hyperparameter optimisation, along with the set of hyperparameters optimised. Brackets indicate the hyperparameter defaults. The size of the separation-set for CI tests is set to -1 to allow for an unlimited size of conditioning sets.

Table 2 lists the five structure learning algorithms considered for hyperparameter optimisation, spanning all three classes of structure learning. Because OTSL is designed to optimise either EBIC$_{normalised\ γ}$ or BDeu$_{iss}$, we follow a somewhat different strategy when optimising constraint-based learning algorithms which do not involve score-based hyperparameters such as iss and γ. As shown in Table 2, the PC-Stable algorithm is tuned by exploring the three different thresholds for significance test α by maximising either EBIC or BDeu given their hyperparameter defaults; i.e., we iterate over hyperparameter values for α – not for iss or γ – when the input algorithm is constraint-based. Specifically, a) for constraint-based PC-stable we optimise hyperparameter α which represents the statistical significance threshold for either Chi$^2$, MI, or MI-sh (refer to Section 2.1), b) for score-based HC and FGES we optimise hyperparameter γ in EBIC$_γ$ and iss in BDeu$_{iss}$ (refer to Section 2.2), and c) for hybrid algorithms MCMC and MMHC we optimise for all three possible hyperparameters. However, as shown in Table 2, we reduce the size of the set of possible hyperparameters to be explored for hybrid algorithms due to the much larger number of possible combinations of hyperparameters they produce. For example, if we were to explore the same range of hyperparameters for MMHC then that would require 1,920 structure learning experiments for that algorithm alone; i.e., 3 hyperparameters for $α \times 20$ for γ or 3 hyperparameters for $α \times 20$ for iss for each case study and sample size.

We use the F1 and SHD graphical metrics to assess synthetic experiments where the ground truth DAG is known. The F1 metric represents the harmonic mean of Precision (P) and Recall (R) where $F1 = 2\frac{P.R}{P+R}$, and the SHD score represents the number of edge additions, deletions and reversals needed to convert the learnt DAG into the true DAG. The scores reported in this study reflect comparisons between learnt and true DAGs. If a structure learning algorithm produces a CPDAG then a random DAG is generated from the learnt CPDAG.

The hyperparameter optimisation performance of OTSL is assessed with reference to other hyperparameter tuning methods that are specifically proposed for tuning structure learning algorithms, and specifically the StARS and OCT approaches discussed in the introduction. In addition, we also consider the BIC and the Akaike Information Criterion (AIC) model-selection functions as baselines for tuning, consistent with how they are used in other relevant studies for evaluation purposes, where tuning is determined by the hyperparameter value that maximises the given model-selection function (Biza et al., 2020).

We conduct all experiments by performing 10 iterations of resampling for both OTSL and StARS, and assuming a 10-fold cross-validation for OCT. We set a runtime limit of 24 hours for each experiment and yet, this was not enough to complete all experiments. Because most tuning experiments failed to complete learning on the real-world Diarrhoea and Weather datasets within the runtime limit, we had to modify the experimental setup for these two datasets. The issue with the Diarrhoea dataset is that it contains a large number of samples (259,627), which we address by modifying the resampling technique such that it creates 10 sets of training data restricted to a sample size of 9k and 10 sets of test data restricted to a sample size of 1k, derived from the 259,627 instances of the Diarrhoea dataset. On the other hand, the issue with the Weather dataset is that it contains a large number of variables, and we address this by reducing the number of iterations for resampling to 5 for the Weather dataset.

Experiments with real data provide no access to ground truth. As a result, it is difficult to judge the unsupervised learning performance of these algorithms on real data. Therefore, we use real data to primarily investigate the issues we may face, specifically with large datasets as discussed above, and to illustrate how OTSL influences the structure learning performance of the different algorithms considered, in terms of model-selection, goodness-of-fit, and model dimensionality.

We test PC-Stable, HC and MMHC using the *bnlearn* R package (Scutari, 2019). FGES using Tetrad-based *rcausal* R package (Wongchokprasitti, 2019), and MCMC (the order-MCMC version) using the *BiDAG* R package (Suter et al., 2023). The model-selection scores of BIC and AIC, as well as the StARS and OCT tuning algorithms are tested using the MATLAB implementations available at: https://github.com/mensxmachina/OCT. The implementation of OTSL is made available online at https://github.com/kiattikunc/OTSL. All experiments were conducted on a high performance computing cluster with 32 GBs of RAM, whereas the experiments involving the FGES algorithm were ran on a laptop with an M1 CPU at 3.2 GHz and 8GB of RAM.

## 5. Empirical results

*5.1 Results based on synthetic data*

*5.1.1 Impact of hyperparameter tuning on graphical structure*

We assume two different cases for hyperparameter defaults: a) *Default A* where $\alpha = 0.05$ for Chi$^2$ test and $\gamma = 0$ for EBIC$_\gamma$, and b) *Default B* where and $\alpha = 0.05$ for Chi$^2$ test and iss $= 1$ for BDeu$_{iss}$. Figure 2 compares the F1 scores obtained by the four specified algorithms across all synthetic experiments, with and without (i.e., *Default A)* hyperparameter optimisation. In this set of experiments, hyperparameter optimisation is restricted to EBIC$_{normalised\ \gamma}$ and hence, the MCMC algorithm is not included in these results since EBIC$_\gamma$ is not available in the *BiDAG* R package. Figure 2a depicts the results when trained with datasets of sample size 1k, whereas Figure 2b depicts the results when trained with datasets of sample size 10k.

Across the 12 comparisons shown in both Figures 2a and 2b, the results show that the hyperparameter tuning applied by OTSL improves the average F1 scores in 9 cases, and slightly decreases performance in 3 cases; i.e., for Property at both 1k and 10k sample sizes and for Sports at 10k sample size. In Figure 2a, the average F1 score across all DAGs learnt over the six cases and four structure learning algorithms is 0.448 for default configurations, and increases to 0.458 (or by ~2.3%) when tuning the hyperparameters of $EBIC_{normalised\ \gamma}$. Figure 2b repeats these experiments for sample sizes 10k and shows that the results remain consistent with those obtained when the sample size is set to 1k. Specifically, the average F1 score across all DAGs is 0.5 for the default configurations, and increases to 0.513 (or by ~2.5%) when tuned with OTSL.

Figures 3a and 3b repeat the experiments of Figures 2a and 2b, and use $BDeu_{iss}$ as the tuning score instead of $EBIC_{normalised\ \gamma}$. In this case, however, the results show that the hyperparameter tuning applied by OTSL did not improve the average F1 scores. Specifically, the average F1 scores for the default configurations (*Default B*) are 0.51 and 0.56 for sample sizes 1k and 10k respectively, and 0.506 and 0.56 respectively when tuned with OTSL.

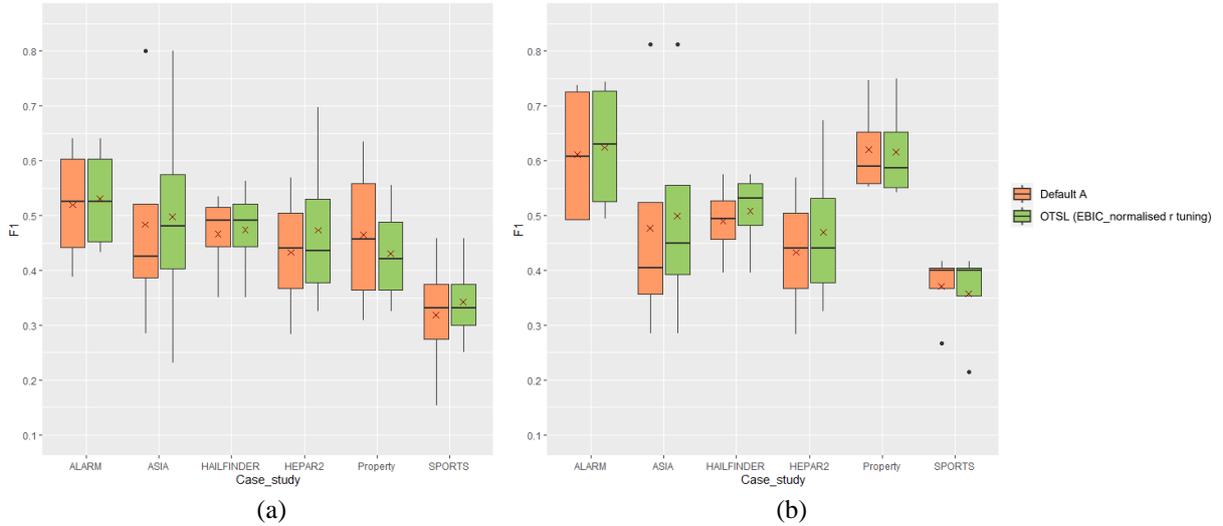

(a)          (b)

**Figure 2**: The average F1 scores with and without hyperparameter tuning. Untuned algorithms assume Default A configuration and tuned algorithms assume OTSL with $EBIC_{normalised\ \gamma}$ as the tuning score. The average scores are derived over four structure learning algorithms (excluding MCMC that does not support $EBIC_\gamma$), and six synthetic case studies. The boxplots represent the highest and lowest F1 scores with outliers, × is the mean and – is the median. The lower edge of the boxplot represents the first quartile, while the higher edge of the boxplot represents the third quartile. Figure (a) depicts the scores for datasets with sample size 1k, and (b) with sample size 10k.

Table 3 details the average change in F1 and SHD scores for each structure learning algorithm relative to the hyperparameter defaults (*Default A)*, when we tune their hyperparameters with OTSL and $EBIC_{normalised\gamma}$ as the tuning score, as well as when we randomise the hyperparameter values averaged over 10 iterations. The results depicted in Table 3 show that randomising the hyperparameters leads to an average *decrease* of 1.71% in F1 score, and a decrease of 4.89% in SHD score, relative to the results obtained when assuming hyperparameter defaults. On the other hand, the F1 and SHD scores *increase* by 3.9% and 6.12% respectively when optimising the hyperparameters using OTSL. However, the constraint-based PC-Stable generates poor tuning performance with F1 and SHD scores decreasing by 1.81% and 1.08% respectively. This might suggest that the score-based tuning applied by OTSL to tune constraint-based CI tests might not be appropriate.

Table 4 repeats the experiments but assumes *Default B* configurations, and that the tuning score is $BDeu_{iss}$ instead of $EBIC_{normalised\ \gamma}$ assumed in Table 3. In this case, the results show that both randomising and optimising the hyperparameter iss of $BDeu_{iss}$ decreases graphical scores relative to those obtained by assuming hyperparameter defaults. In other words, it seems that assuming iss = 1 for $BDeu_{iss}$ produces strong performance with little, if any, room for improvement via hyperparameter tuning, and this is consistent with what is reported by Steck (2008) and Uneo (2011) who recommend to set iss = 1, especially when the distributions of the variables are assumed to be skewed or when the true underlying structure is assumed to be sparse. Our results show that randomising the iss hyperparameter of $BDeu_{iss}$ *decreases* the F1 and SHD scores by 4.86% and 11.02%, whereas optimising iss with OTSL *increases* the F1 scores by 0.12% and *decreases* the SHD scores by 3.36%. These results suggest that the $BDeu_{iss}$ function may not be suitable for tuning, at least compared to $EBIC_{normalised\ \gamma}$, and that setting iss = 1 might indeed be sufficient, in general.

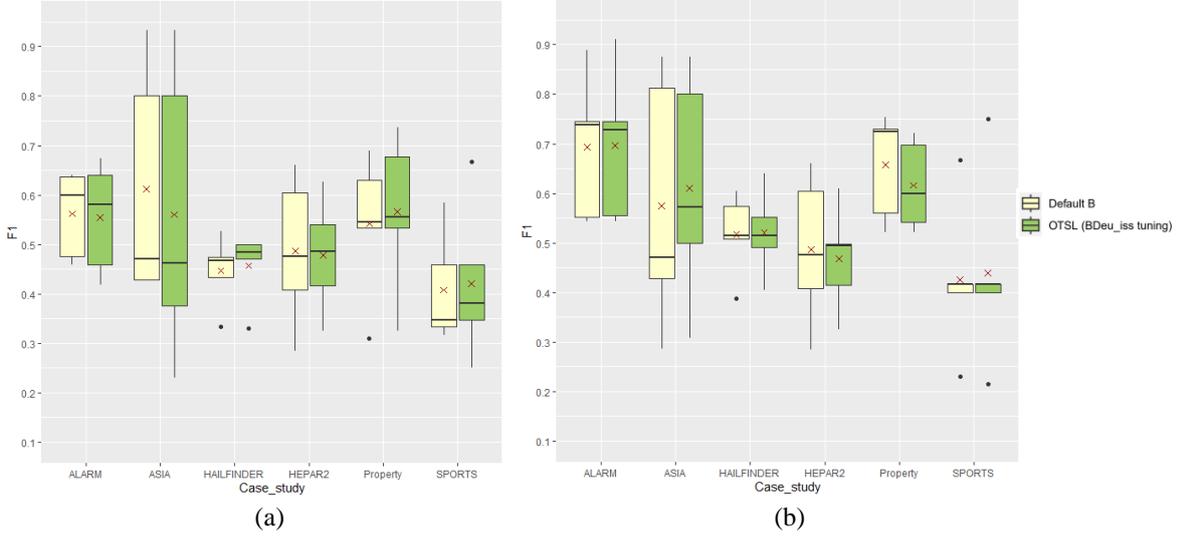

(a)          (b)

**Figure 3**: The average F1 scores with and without hyperparameter tuning. Untuned algorithms assume Default B configuration and tuned algorithms assume OTSL with $BDeu_{iss}$ as the tuning score. The average scores are derived over five structure learning algorithms, and six synthetic case studies. The boxplots represent the highest and lowest F1 scores with outliers, × is the mean and – is the median. The lower edge of the boxplot represents the first quartile, while the higher edge of the boxplot represents the third quartile. Figure (a) depicts the scores for datasets with sample size 1k, and (b) with sample size 10k.

| Algorithm | Change in F1 relative to Default A | | Change in SHD relative to Default A | |
|---|---|---|---|---|
| | Random configuration | Tuning with $EBIC_{normalised\ \gamma}$ | Random configuration | Tuning with $EBIC_{normalised\ \gamma}$ |
| PC-Stable | **0.95%** | -1.18% | -1.30% | -1.08% |
| HC | 6.29% | **12.96%** | 8.29% | **23.40%** |
| FGES | -14.21% | -0.10% | -26.31% | **1.54%** |
| MMHC | 0.00% | **3.91%** | -0.43% | **0.61%** |
| Average | -1.71% | **3.90%** | -4.89% | **6.12%** |

**Table 3**: The change in average F1 and SHD scores for each algorithm, after randomising their hyperparameters and after tuning them with OTSL. The experiments consider all six synthetic case studies and both 1k and 10k sample sizes. The hyperparameter defaults are $\alpha = 0.05$ for $Chi^2$ test and $\gamma = 0$ for $EBIC_\gamma$ (*Default A*). The best performance values are shown in bold.

| Algorithm | Change in F1 relative to Default B | | Change in SHD relative to Default B | |
|---|---|---|---|---|
| | Random configuration | Tuning with $BDeu_{iss}$ | Random configuration | Tuning with $BDeu_{iss}$ |
| PC-Stable | 0.78% | **2.40%** | -2.00% | **0.64%** |
| HC | -10.48% | -8.64% | -18.95% | -11.33% |
| FGES | -14.77% | **2.25%** | -34.65% | -9.68% |
| MCMC | **2.79%** | 2.51% | 2.40% | **3.25%** |
| MMHC | -2.69% | **2.06%** | -2.11% | **0.31%** |
| Average | -4.86% | **0.12%** | -11.02% | -3.36% |

**Table 4**: The change in average F1 and SHD scores for each algorithm, after randomising their hyperparameters and after tuning them with OTSL. The experiments consider all six synthetic case studies and both 1k and 10k sample sizes. The hyperparameter defaults are $\alpha = 0.05$ for $Chi^2$ test and $\gamma = 0$ for $BDeu_{iss}$ (*Default B*). The best performance values are shown in bold.

*5.1.2. Assessing OTSL relative to existing tuning algorithms for structure learning*

We compare the results of OTSL with those obtained by the out-of-sample tuning OCT and the in-sample tuning StARS. We also consider the baseline tuning results obtained by the model-selection scores BIC and AIC. This process involves applying the other four approaches to the same experiments presented in subsection 5.1.1, and comparing the changes to the F1 and SHD scores across all hyperparameter tuning approaches.

| Structure learning algorithm | Hyperparameter tuning method | | | | | | | | | |
|---|---|---|---|---|---|---|---|---|---|---|
| | Out-of-sample | | In-sample | | | Out-of-sample | | In-sample | | |
| | OTSL with $EBIC_{normalised\,\gamma}$ tuning | OCT | Model selection with BIC | Model selection with AIC | StARS | OTSL with $EBIC_{normalised\,\gamma}$ tuning | OCT | Model selection with BIC | Model selection with AIC | StARS |
| | Change of F1 relative to Default A | | | | | Change of SHD relative to Default A | | | | |
| PC-Stable | -1.18% | **1.88%** | 4.32% | 3.30% | 0.72% | -1.08% | **0.09%** | 1.21% | 1.56% | -0.25% |
| HC | **12.96%** | 6.23% | -0.65% | -1.60% | -3.62% | **23.40%** | -35.43% | -8.76% | -9.90% | -34.77% |
| FGES | -0.10% | **1.26%** | -6.87% | -7.43% | 1.55% | **1.54%** | -0.67% | -5.71% | -7.37% | -2.59% |
| MMHC | 3.91% | 3.03% | 11.55% | **19.88%** | 3.36% | 0.61% | -4.89% | 22.47% | **23.08%** | 17.09% |
| Average | **3.90%** | 3.10% | 2.09% | 3.54% | 0.50% | **6.12%** | -10.22% | 2.30% | 1.84% | -5.13% |

**Table 5**: The average change in F1 and SHD scores due to hyperparameter tuning by the specified tuning method. The averages are derived from all six synthetic case studies and over both sample sizes. The structure learning algorithms assume *Default A* hyperparameter configuration (Chi$^2$ test with $\alpha = 0.05$, and $EBIC_\gamma$ with $\gamma = 0$). The highest improvements in graphical accuracy are shown in bold.

Tables 5 and 6 summarise these results for both *Default A* and *Default B* hyperparameter configurations respectively. Table 5 shows that while none of the hyperparameter tuning approaches improves the graphical accuracy for all four structure learning algorithms, most of the approaches do improve the average structure learning performance across all algorithms. Specifically, all five tuning approaches improve the average F1 score across the four structure learning algorithms considered, although only three out of the five tuning approaches also improve the SHD score. The OTSL algorithm with $EBIC_{normalised\,\gamma}$ tuning increases both the F1 (up by 3.9%) and SHD (up by 6.12%) scores the most across all the tuning approaches considered. Interestingly, the F1 and SHD scores provide contradictory conclusions about the impact on graphical structure for OCT and StARS algorithms, and this inconsistency between the F1 and SHD metrics is in agreement with other studies (Constantinou et al., 2021). For example, the F1 metric suggests that the hyperparameter tuning of OCT improves the structure learning performance of all four structure learning algorithms, whereas the SHD metric suggests that OCT decreases the graphical accuracy of three out of the four structure learning algorithms.

| Structure learning algorithm | Hyperparameter tuning method | | | | | | | | | |
|---|---|---|---|---|---|---|---|---|---|---|
| | Out-of-sample | | In-sample | | | Out-of-sample | | In-sample | | |
| | OTSL with $BDeu_{iss}$ tuning | OCT | Model selection with BIC | Model selection with AIC | StARS | OTSL with $BDeu_{iss}$ tuning | OCT | Model selection with BIC | Model selection with AIC | StARS |
| | Change of F1 relative to Default B | | | | | Change of SHD relative to Default B | | | | |
| PC-Stable | **2.40%** | 1.26% | -6.87% | -7.43% | 1.55% | **0.64%** | -0.67% | -5.71% | -7.37% | -2.59% |
| HC | -8.64% | -11.25% | -4.26% | **1.88%** | -11.21% | -11.33% | -40.02% | -2.57% | **-2.06%** | -11.15% |
| FGES | **2.25%** | -0.43% | -6.02% | -6.94% | -7.91% | **-9.68%** | -41.77% | -15.96% | -17.17% | -42.71% |
| MCMC | **2.51%** | -2.39% | -0.93% | 0.18% | -0.15% | **3.25%** | -10.80% | -7.93% | 1.97% | -2.29% |
| MMHC | **2.06%** | -2.47% | 0.41% | -0.59% | -3.56% | **0.31%** | -1.65% | -0.60% | -0.28% | -2.09% |
| Average | **0.12%** | -3.06% | -3.53% | -2.58% | -4.26% | **-3.36%** | -18.98% | -6.55% | -4.98% | -12.17% |

**Table 6**: The average change in F1 and SHD scores due to hyperparameter tuning by the specified tuning method. The averages are derived from all six synthetic case studies and over both sample sizes. The structure learning algorithms assume *Default B* hyperparameter configuration (Chi$^2$ test with $\alpha = 0.05$, and $BDeu_{iss}$ with iss = 1). The highest improvements in graphical accuracy are shown in bold.

Table 6 presents the same results when the hyperparameter tuning approaches are applied to the iss hyperparameter of $BDeu_{iss}$. Overall, the results are consistent with those presented in Tables 3 and 4, in that hyperparameter tuning appears to be successful for $EBIC_{normalised\,\gamma}$ but not for $BDeu_{iss}$.

While tuning with $BDeu_{iss}$ is found to be rather inadequate for all tuning methods, OTSL is found to perform considerably better compared to the other tuning approaches with an *increase* of 0.12% in the average F1 score (improved the scores of four out of the five algorithms) and a *decrease* of 3.36% in the average SHD score (improved the scores of three out of the five algorithms).

We also assess the computational complexity of OTSL by comparing its hyperparameter tuning and overall structure learning runtimes against those produced by the other hyperparameter tuning approaches. Provisional results show that the runtimes are similar for both $EBIC_{normalised\gamma}$ and $BDeu_{iss}$, but here we focus on $EBIC_{normalised\ \gamma}$ which produces the best tuning performance. Figure 4a depicts the total runtimes (hyperparameter tuning and structure learning) across all six case studies, two sample sizes, and five structure learning algorithms, whereas Figure 4b shows the runtime for the same experiments but restricted to the hyperparameter tuning phase. As expected, optimisation with model-selection functions such as BIC and AIC results in very low runtimes, since they do not involve out-of-sample or resampling strategies, whereas OTSL, OCT and StARS perform 10 iterations of either in-sample or out-of-sample tuning for each hyperparameter configuration and hence, they produce considerably higher runtimes. Overall, the results in Figure 4a show that the computational runtime of OTSL is similar to that of StARS, and considerably faster than that of OCT. Importantly, the tuning runtimes of OTSL and StARS represent just 0.2% and 0.4% of the total structure learning runtime respectively, whereas the tuning runtime of OCT represents 43% of its total structure learning runtime. Figure 4b shows that the tuning runtime of OTSL is slower than the tuning runtime of StARS, but much faster than the tuning runtime of OCT.

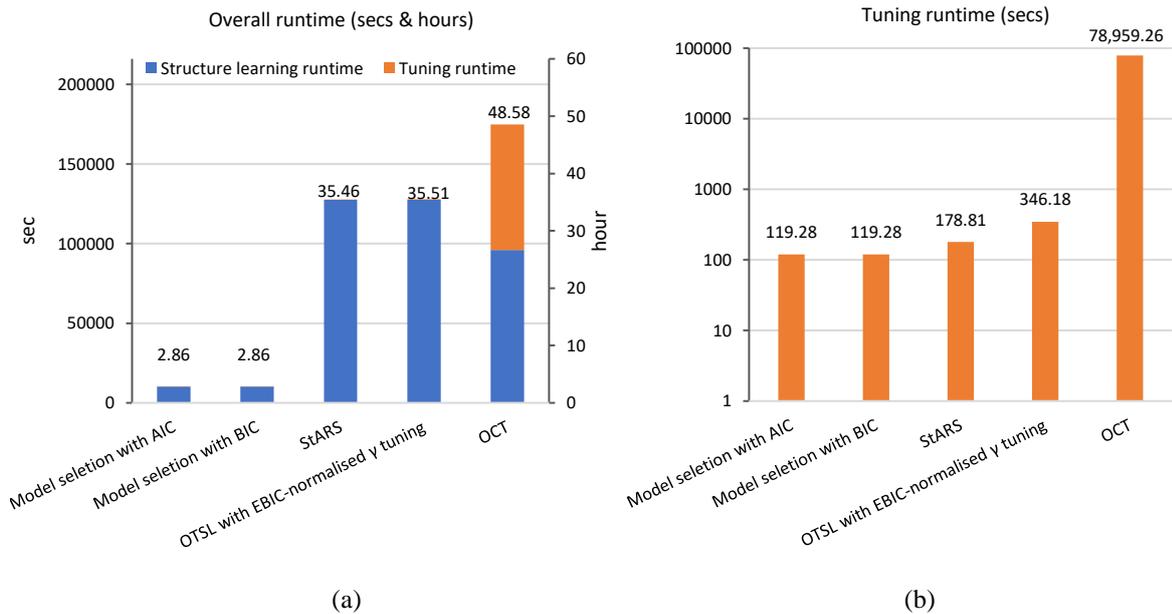

(a) (b)

**Figure 4**: (a) Overall runtime (structure learning and tuning) and (b) tuning runtime, summed over all six synthetic datasets and two sample sizes, across all five structure learning algorithms.

### 5.2. Applying OTSL to real data

While previous subsections focused on evaluating OSTL in terms of how its tuning improves the recovery of the ground truth graphs that were used to generate synthetic data, this subsection illustrates how OTSL could be used in practice with application to four different real datasets that come from different disciplines. As discussed in Section 3, real data do not come with an access to ground truth and hence, the purpose here is to illustrate how tuning influences the structure learning performance of the different algorithms considered when applied to real data. We consider the following four discrete datasets, where the first three are obtained from the Bayesys repository (Constantinou et al., 2020) and the fourth from the National Center for Environmental Prediction (NCEP) and the National Center for Atmospheric Research (NCAR) in the USA, known as the NCEP/NCAR Reanalysis Project (Kalnay et al., 1996):

a) **ForMed**: A case study on assessing and managing the risk of violence in released prisoners with history of violence and mental health (Coid et al., 2016; Constantinou et al., 2015). The data was collected through interviews and assessments comprising risk factors for 953 individual cases. The dataset contains a total of 56 categorical variables.

b) **Covid-19**: A dataset that captures pandemic data about the COVID-19 outbreak in the UK (Constantinou et al., 2023). The data comprises of 18 variables that capture information related to viral tests, infections, hospitalisations, vaccinations, deaths, COVID-19 variants, population mobility such as usage of transportation, schools, and restaurants, as well as various government policies such as facemasks and lockdowns. The data instances represent daily information, spanning from January 30$^{th}$, 2020, to June 13$^{th}$, 2022, resulting in a total of 866 instances.

c) **Diarrhoea**: Survey data collated and pre-processed from the Demographic and Health Survey (DHS) program, which was used to investigate the factors associated with childhood diarrhoea in India (Kitson and Constantinou, 2021). The dataset captures relevant cases from 2015 to 2016 and contains 28 variables and 259,627 instances.

d) **Weather**: A dataset that captures the monthly means of air temperature and other climatological data for each location as measured by latitude (y coordinate) and longitude (x coordinate) over the global grid system (Kalnay et al., 1996). The dataset merges information obtained from multiple sources, i.e., balloons, satellites, and buoys. It provides a comprehensive 75-year record from 1948 to 2022 of global atmospheric field analyses. We used the *bnlearn* R package (2019) to discretise the dataset. Because the raw data is too big for our experiments, we also resized the spatial dataset from 2.5 degree x 2.5 degree global grids to 10 degree x 10 degree global grids, and reduced the total number of variables from 10,512 (144x73) to 648 (36x18). Therefore, the dataset used in this study contains a total of 648 variables and 900 instances.

We apply the structure learning algorithms to each of the four datasets, and tune their hyperparameters using OTSL. We only consider *Default A* hyperparameter configuration with $EBIC_{normalised\ \gamma}$ for tuning, which was shown to be more suitable for hyperparameter optimisation. Table 7 presents the results obtained by applying the specified structure learning algorithms to the ForMed dataset and tuning their hyperparameters with OTSL. We report the model-selection score BIC, the goodness-of-fit score LL, the number of free parameters as a measure of model dimensionality, and the tuning scores $EBIC_{normalised\ \gamma}$. Table 7 shows that out of the four structure learning algorithms considered, only one (HC with $\gamma = 2$) had its hyperparameter changed following tuning with OTSL. The tuning scores $EBIC_{normalised\ \gamma}$ in Table 7 suggest that the graph produced by MMHC, presented in Figure 5, might be the 'best' structure to consider amongst those learnt by the different algorithms, although this suggestion contradicts the BIC score which suggests that the best structure may be the one learnt by HC.

Tables 8, 9, and 10, and corresponding Figures 6, 7, and 8, repeat the above analyses for case studies Covid-19, Diarrhoea and Weather respectively. Note that only two algorithms are reported for the Weather case study, and this is because HC did not complete learning within the 24-hour time limit, while FGES returned a memory allocation error. The results show that OTSL modified the hyperparameters of three and four, out of the four, structure learning algorithms in Covid-19 and Diarrhoea cases respectively, and for one out of the two algorithms for dataset Weather. Table 8 shows that FGES produces the best structure for the Covid-19 case study (see Figure 6) according to both $EBIC_{normalised\ \gamma}$ and BIC scores. On the other hand, the results in Table 9 suggest that the graph

produced by FGES is the best structure according to $EBIC_{normalised\ \gamma}$, which once more contradicts the BIC score that scores the graph produced by HC the highest. Lastly, in Table 10 both $EBIC_{normalised\ \gamma}$ and BIC are in agreement that MMHC produced the best structure shown in Figure 8. The nodes in Figure 8 represent random variables of a monthly temperature for each location, whereas the arcs represent the spatial dependencies of surface temperatures for each grid[1].

| Structure learning algorithm | CI test / Objective score | Optimal hyperparameter from OTSL (Tuning with $EBIC_{normalised\ \gamma}$) | Tuning score $EBIC_{normalised\ \gamma}$ from OTSL | Score of learnt graph BIC | LL | Free parameters |
|---|---|---|---|---|---|---|
| PC-Stable | $Chi^2$ | $\alpha = 0.05$ | -4,099 | -40,791 | -39,775 | 296 |
|  | MI | $\alpha = 0.05$ | -4,113 | -40,799 | -39,760 | 303 |
|  | Mi-sh | $\alpha = 0.05$ | -4,092 | -40,837 | -39,774 | 310 |
| HC | $EBIC_\gamma$ | $\gamma = 2$ | -3,974 | **-37,062** | -35,442 | 540 |
| FGES | $EBIC_\gamma$ | $\gamma = 0$ | -4,195 | -42,343 | -41,846 | 145 |
| **MMHC** | **$Chi^2$ / $EBIC_\gamma$** | **$\alpha = 0.05 / \gamma = 0$** | **-3,942** | -38,183 | -37,744 | 439 |

**Table 7**: The tuning, model-selection, goodness-of-fit, and dimensionality scores of the graphs learnt by the specified structure learning algorithms when applied to the ForMed dataset, with OTSL tuning. The best performance values are shown in bold.

| Structure learning algorithm | CI test / Objective score | Optimal hyperparameter from OTSL (Tuning with $EBIC_{normalised\ \gamma}$) | Tuning score $EBIC_{normalised\ \gamma}$ from OTSL | Score of learnt graph BIC | LL | Free parameters |
|---|---|---|---|---|---|---|
| PC-Stable | $Chi^2$ | $\alpha = 0.1$ | -1,392 | -13,666 | -13,270 | 117 |
|  | MI | $\alpha = 0.1$ | -1,395 | -13,768 | -13,190 | 171 |
|  | Mi-sh | $\alpha = 0.01$ | -1,395 | -13,768 | -13,190 | 171 |
| HC | $EBIC_\gamma$ | $\gamma = 5$ | -1,249 | -10,725 | -8,787 | 323 |
| **FGES** | **$EBIC_\gamma$** | **$\gamma = 1$** | **-1,092** | **-9,038** | -9,918 | 260 |
| MMHC | $Chi^2$ / $EBIC_\gamma$ | $\alpha = 0.05 / \gamma = 0$ | -1,257 | -12,267 | -12,129 | 138 |

**Table 8**: The tuning, model-selection, goodness-of-fit, and dimensionality scores of the graphs learnt by the specified structure learning algorithms when applied to the COVID-19 dataset, with OTSL tuning. The best performance values are shown in bold.

---

[1] The orange arcs represent short-distance temperature dependencies, while the red arcs show the teleconnected dependencies. We observe that the local short-distance arcs are dense, representing atmospheric thermodynamic processes, while the teleconnected dependencies are represented by only three arcs. One of these teleconnected dependencies indicates the El Niño effects, which are caused by temperatures along the equator in the Pacific Ocean (Yamasaki et al., 2008).

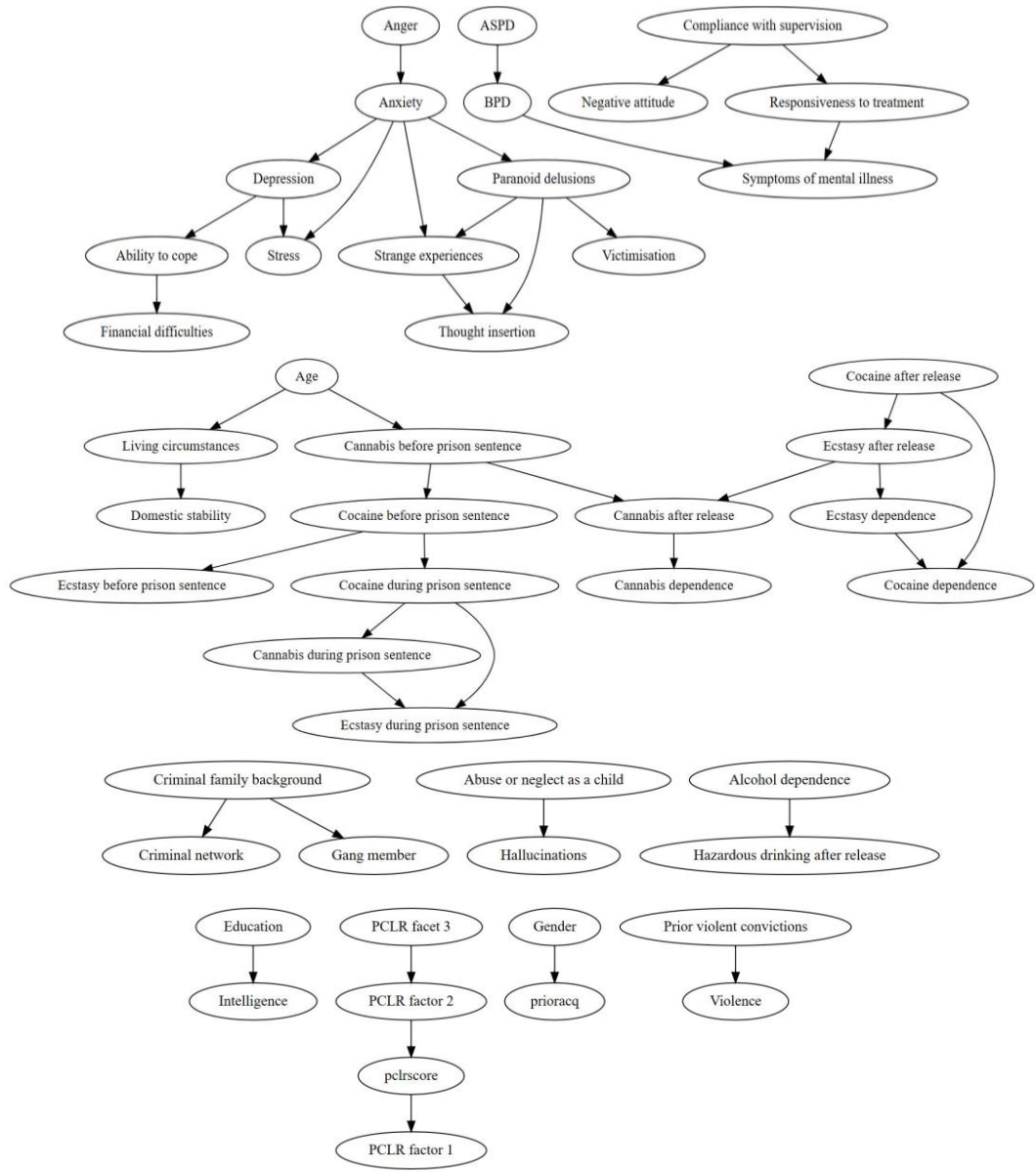

**Figure 5**: The DAG learnt by MMHC for the ForMed dataset with OTSL tuning (Table 7).

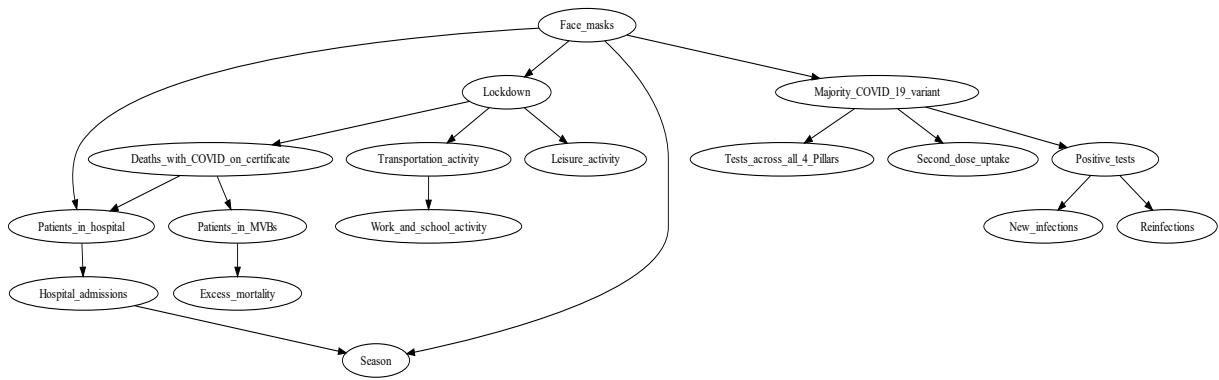

**Figure 6**: The DAG (sampled from the learnt CPDAG) learnt by FGES for the COVID-19 dataset with OTSL tuning (Table 8).

| Structure learning algorithm | CI test / Objective score | Optimal hyperparameter from OTSL (Tuning with EBIC$_{normalised\ \gamma}$) | Tuning score EBIC$_{normalised\ \gamma}$ from OTSL | Score of learnt graph | | |
|---|---|---|---|---|---|---|
| | | | | BIC | LL | Free parameters |
| PC-Stable | Chi$^2$ | $\alpha = 0.01$ | -19,653 | -4,910,813 | -4,899,630 | 1,794 |
| | MI | $\alpha = 0.1$ | -19,400 | -5,099,129 | -5,088,588 | 1,691 |
| | Mi-sh | $\alpha = 0.01$ | -19,506 | -5,082,642 | -5,068,485 | 1,691 |
| HC | EBIC$_\gamma$ | $\gamma = 2$ | -19,257 | **-4,776,526** | -4,748,359 | 9,389 |
| **FGES** | **EBIC$_\gamma$** | **$\gamma = 0$** | **-19,175** | -4,944,463 | -4,941,340 | 501 |
| MMHC | Chi$^2$ / EBIC$_\gamma$ | $\alpha = 0.01$ / $\gamma = 0$ | -19,257 | -4,979,854 | -4,979,334 | 520 |

**Table 9**: The tuning, model-selection, goodness-of-fit, and dimensionality scores of the graphs learnt by the specified structure learning algorithms when applied to the Diarrhoea dataset, with OTSL tuning. The best performance values are shown in bold.

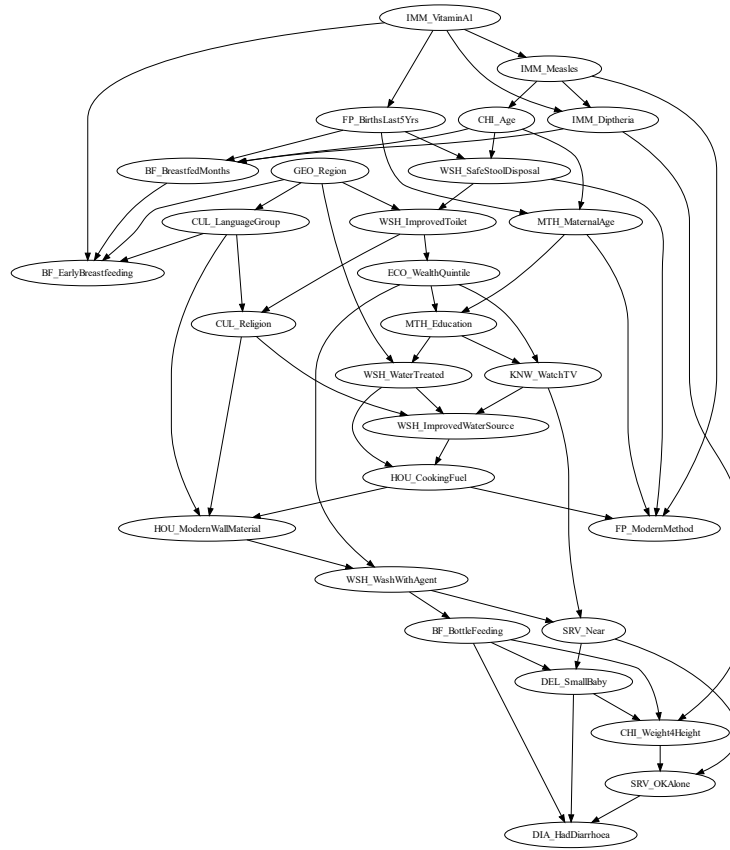

**Figure 7**: The DAG (sampled from the learnt CPDAG) learnt by FGES for the Diarrhoea dataset with OTSL tuning (Table 9).

| Structure learning algorithm | CI test / Objective score | Optimal hyperparameter from OTSL (Tuning with EBIC$_{normalised\ \gamma}$) | Tuning score EBIC$_{normalised\ \gamma}$ from OTSL | Score of learnt graph | | |
|---|---|---|---|---|---|---|
| | | | | BIC | LL | Free parameters |
| PC-Stable | Chi$^2$ | $\alpha = 0.05$ | -67,594 | -334,171 | -319,219 | 4,396 |
| | MI | $\alpha = 0.01$ | -72,525 | -378,849 | -366,591 | 3,604 |
| | Mi-sh | $\alpha = 0.1$ | -71,653 | -346,160 | -333,059 | 3,852 |
| **MMHC** | **Chi$^2$ / EBIC$_\gamma$** | **$\alpha = 0.05, \gamma = 0$** | **-66,350** | **-318,220** | **-314,724** | **3,496** |

**Table 10**: The tuning, model-selection, goodness-of-fit, and dimensionality scores of the graphs learnt by the specified structure learning algorithms when applied to the Weather dataset, with OTSL tuning. The best performance values are shown in bold.

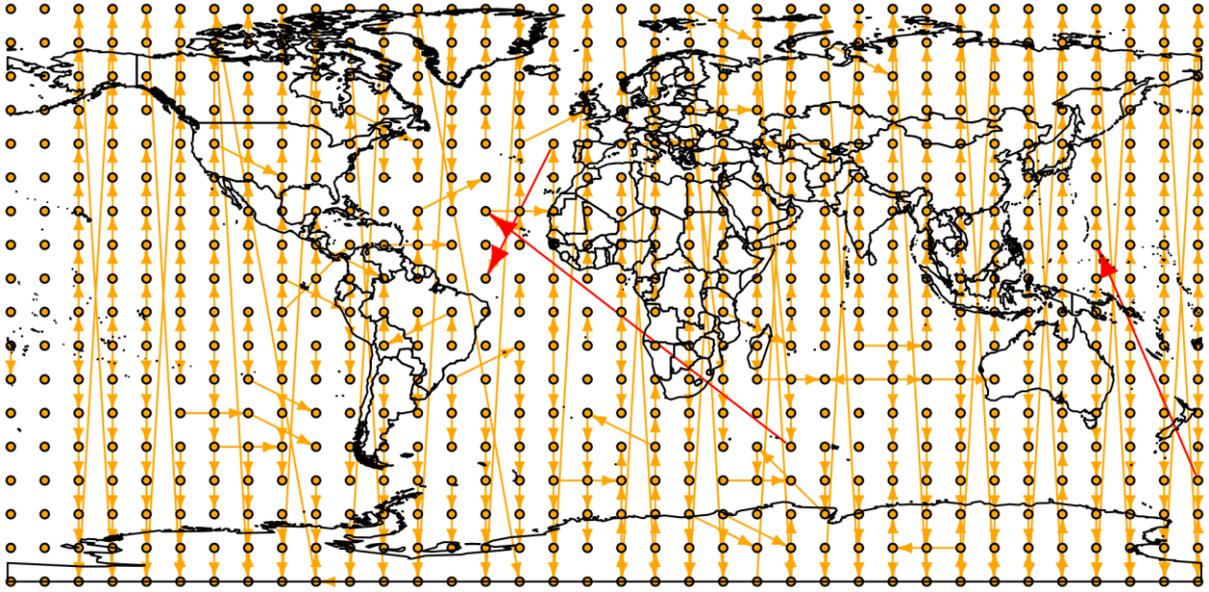

**Figure 8**: The DAG learnt by MMHC for the Weather dataset with OTSL tuning (Table 10). The vertices of the world map superimposed over the DAG represent latitude and longitude locations on 10x10 degree grids.

## 6. Conclusions and future work

Learning causal models from observational data remains a major challenge. Traditionally, structure learning algorithms are evaluated and applied to real data with their hyperparameter defaults, or by iterating over a small set of possible hyperparameters. However, no specific set of hyperparameters is optimal for all input datasets which vary in sample size and dimensionality, and structure learning algorithms which vary in learning strategy. Therefore, the question of which hyperparameter values might be best for a given structure learning algorithm and input dataset combination remains an open question.

In this study, we propose and evaluate a hyperparameter tuning algorithm, called OTSL, that employs out-of-sample resampling and score-based tuning to find the optimal hyperparameters for a given structure learning algorithm, given the input data. We describe and implement OTSL with a focus on score-based learning, and determine the hyperparameters of different algorithms by optimising either the iss or γ hyperparameters of $BDeu_{iss}$ and $EBIC_{normalised\ \gamma}$ objective scores.

Synthetic experiments show that tuning with OTSL leads to reasonable improvements in structure learning in terms of the F1 and SHD scores, and when assuming $EBIC_\gamma$ as the objective score for score-based learning. However, this level of improvement is not repeated for $BDeu_{iss}$, and this observation is consistent for OTSL and all the other tuning approaches investigated in this study. This is because the hyperparameter default of iss = 1 in $BDeu_{iss}$ tends to lead to higher F1 and SHD scores compared to the graphs learnt when iss > 1 (and hence benefits little, if any, from hyperparameter tuning), and this observation is consistent with past studies (Steck, 2008; Uneo, 2011).

The tuning performance of OTSL is evaluated with reference to other hyperparameter tuning approaches for structure learning. We have considered the OCT and StARS tuning approaches, as well as the BIC and AIC model-selection scores that serve as baselines for tuning hyperparameters. Overall, the results show that OTSL provides better tuning performance from results derived across different structure learning algorithms, case studies, and sample sizes. In terms of computational complexity, OTSL was found to be more efficient than OCT but slightly less efficient than StARS.

A limitation is that while OTSL can be applied to structure learning algorithms that come from different classes of learning, it is designed with score-based learning in mind and assumes that the optimal hyperparameters are those that maximise either the $\text{EBIC}_{\text{normalised }\gamma}$ or $\text{BDeu}_{\text{iss}}$ objective scores, and this also applies when tuning CI functions in constraint-based learning. This might explain why the results from tuning score-based learning algorithms are better that those derived from tuning constraint-based learning. Another limitation is that, because OTSL optimises hyperparameters on test data, this process involves resampling multiple training and test datasets from a single input dataset, which impacts the computational efficiency of structure learning; a learning process that is already known to be computationally expensive.

**Acknowledgements**

This research was supported by the Royal Thai Government Scholarship offered by Thailand's Office of Civil Service Commission (OCSC).